\documentclass[letterpaper, 10 pt, journal, twoside]{IEEEtran}

\IEEEoverridecommandlockouts                              % This command is only needed if 
\title{\approach: Motion-Aware Fast and Robust Camera Localization for Dynamic Neural Radiance Fields
}

\author{Nicolas Schischka$^{1*}$, Hannah Schieber$^{2,3*}$, Mert Asim Karaoglu$^{1,4*}$, Melih Gorgulu$^{1}$, Florian Grötzner$^{1}$, \\ Alexander Ladikos$^{4}$, Nassir Navab$^{1,5}$,  Daniel Roth$^{2}$, and Benjamin Busam$^{1}$ %  <-this % stops a space
\thanks{Manuscript received: August, 14, 2024; Revised October, 28, 2024; Accepted November, 21, 2024.}%Use only for final RAL version
\thanks{This paper was recommended for publication by Editor Javier Civera upon evaluation of the Associate Editor and Reviewers' comments.
We thank d.hip for providing a campus stipend.}
\thanks{$*$equal contribution}%
\thanks{$^{1}$Mert Asim Karaoglu, Benjamin Busam, Nassir Navab, Nicolas Schischka, Melih Gorgulu, Florian Grötzner are with Technical University of Munich, Munich, Germany {\tt\small mert.karaoglu@tum.de}}%
\thanks{$^{2}$Daniel Roth and Hannah Schieber are with Human-Centered Computing and Extended Reality Lab, Technical University of Munich, School of Medicine and Health, Klinikum rechts der Isar, Orthopedics and Sports Orthopedics, Munich, Germany {\tt\small hannah.schieber@tum.de}}% 
\thanks{$^{3}$Hannah Schieber is with the Friedrich-Alexander Universität Erlangen-Nürnberg, Erlangen, Germany}%
\thanks{$^{4}$Mert Asim Karaoglu and Alexander Ladikos are with ImFusion GmbH, Munich, Germany}%
\thanks{$^{5}$Nassir Navab is with Johns Hopkins University, Baltimore, MD, USA}%
\thanks{Digital Object Identifier (DOI): \textit{10.1109/LRA.2024.3511399}.}
}

\newcommand{\specialcell}[2][l]{%
\begin{tabular}[#1]{@{}c@{}}#2\end{tabular}
}
\usepackage{multirow}
\usepackage{graphicx}
\usepackage{amsmath} % assumes amsmath package installed
\usepackage[bb=dsserif]{mathalpha}
\usepackage{xspace}
\def\approach{DynaMoN\xspace}

\begin{document}

\maketitle
%\thispagestyle{empty}
%\pagestyle{empty}

%%%%%%%%%%%%%%%%%%%%%%%%%%%%%%%%%%%%%%%%%%%%%%%%%%%%%%%%%%%%%%%%%%%%%%%%%%%%%%%%
\begin{abstract}

The accurate reconstruction of dynamic scenes with neural radiance fields is significantly dependent on the estimation of camera poses.
Widely used structure-from-motion pipelines encounter difficulties in accurately tracking the camera trajectory when faced with separate dynamics of the scene content and the camera movement.
To address this challenge, we propose \underline{Dyna}mic \underline{Mo}tion-Aware Fast and Robust Camera Localization for Dynamic \underline{N}eural Radiance Fields (\approach). 
\approach utilizes semantic segmentation and generic motion masks to handle dynamic content for initial camera pose estimation and statics-focused ray sampling for fast and accurate novel-view synthesis.
Our novel iterative learning scheme switches between training the NeRF and updating the pose parameters for an improved reconstruction and trajectory estimation quality.
The proposed pipeline shows significant acceleration of the training process.
We extensively evaluate our approach on two real-world dynamic datasets, the TUM RGB-D dataset and the BONN RGB-D Dynamic dataset. \approach improves over the state-of-the-art both in terms of reconstruction quality and trajectory accuracy.
We plan to make our code public to enhance research in this area. Code available:  \textit{https://hannahhaensen.github.io/DynaMoN/}.

\end{abstract}

%%%%%%%%%%%%%%%%%%%%%%%%%%%%%%%%%%%%%%%%%%%%%%%%%%%%%%%%%%%%%%%%%%%%%%%%%%%%%%%%

\section{INTRODUCTION}

\IEEEPARstart{E}{nabling} novel view synthesis (NVS) on dynamic scenes often requires multi-camera setups~\cite{broxton2020Immersive}. However, everyday dynamic scenes are often captured by one single moving camera, restricting the field of view~\cite{broxton2020Immersive}. To receive NVS or a 3D/4D scene representation, accurate camera poses are essential for such a representation. One way to retrieve camera poses is  Simultaneous Localization and Mapping (SLAM). SLAM can result in multiple scene representations such as explicit voxel or surfel representations or neural radiance fields (NeRF) representations.
The latter allows the rendering of photorealistic novel views from new, reasonable camera positions~\cite{rosinol2022nerf,muller2022instant}. While many are restricted to static scenes, dynamic NeRF~\cite{cao2023hexplane,shao2023tensor4d,song2023nerfplayer,TiNeuVox} allows NVS for dynamic scenes. 

\begin{figure}[t]
    \centering
     % \framebox{\parbox{3in}{
     \includegraphics[width=\columnwidth]{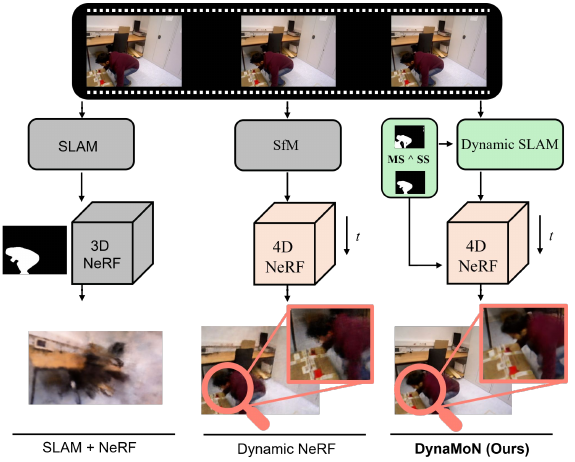}    
    \caption{NeRF combined with SLAM usually relies on static scenes (left). Approaches such as InstantNGP~\cite{muller2022instant} can be used to mask out dynamic content~\cite{rosinol2022nerf,chung2023orbeez} (left) resulting still in reduced quality. Fully dynamic NeRF provides an implicit 4D (3D+time) representation; however, it relies on offline SfM, which can suffer in the presence of considerable motion (center). \approach considers dynamics at all stages using motion segmentation (MS) and semantic segmentation (SS) (right). This enables a more robust camera tracking and NVS with a higher quality (ours, right).}
    \label{fig:intro}
\end{figure}

To enable NVS, accurate camera poses, usually retrieved via structure-from-motion (SfM)~\cite{schonberger2016structure} or the Apple ARKit, are essential. SfM often demands hours of computation to estimate reasonable camera poses and is challenged by large-scale dynamic scenes.
To overcome the limitation of time-intensive SfM, recent works combined SLAM and NeRF~\cite{zhu2022nice, zhu2023nicer,chung2023orbeez,rosinol2022nerf}. SLAM provides faster results for the camera trajectory than classic SfM. However, by nature, scenes are more often dynamic than static. Existing Radiance Field-SLAM~\cite{zhu2022nice, zhu2023nicer,chung2023orbeez,rosinol2022nerf,Matsuki_2024_CVPR,zhu2024semgaussslamdensesemanticgaussian} methods assume a static scene which poses difficulties when dynamic scene content is present as shown in Fig~\ref{fig:intro}. 

In this paper, we present \approach, a motion-aware camera localization and visualization approach that can handle highly dynamic scenes.

\begin{enumerate}
    \item We propose a novel combination of semantic segmentation and generic motion masks during our camera localization to enable robust tracking of the camera path in a dynamic environment for dynamic neural radiance fields.
    \item A statics-focused ray sampling building upon our dynamic masks is introduced to enhance our dynamic NeRF representation by refining the camera poses.
\end{enumerate}

Overall, (1) and (2) improve the camera poses and in turn improve the NVS, leading to improved NVS quality in scenes with strong motions.

We conduct experiments on two challenging datasets, namely the dynamic subset of the TUM RGB-D~\cite{sturm12iros} and the BONN RGB-D Dynamic~\cite{palazzolo2019iros} dataset.
The considered masking in \approach allows us to outperform existing approaches in terms of trajectory error and NVS results on these datasets. 

\section{RELATED WORK}

\approach combines dynamic NeRF with fast and robust motion-aware camera localization. Therefore,  related work covers camera localization, camera localization for NVS, and neural representation for dynamic scenes.

\subsection{Camera Localization and Scene Dynamics}

Previous work on camera localization based on visual odometry (VO)~\cite{engel2017direct}, SfM~\cite{schonberger2016structure} and SLAM~\cite{mur2015orb,campos2021orb,teed2021droid,mur2017orb} extensively study architectures built for static scene assumptions.
However, these methods show limited performance and often fail when they are encountered with the dynamic content commonly existent in real-world scenes.

In recent years, various classical and learning-based methods have been introduced to induce and handle the dynamic motion of the scene for robust camera pose estimation.
With an assumption of prior knowledge of the objects that can independently move in the scene, DS-SLAM~\cite{yu2018ds} builds upon ORB-SLAM2~\cite{mur2017orb} and integrates a semantic segmentation network to mask out motion-prone content along with a moving consistency check in the tracking process.
Similarly, Dai et al.~\cite{dai2020rgb} extend ORB-SLAM2~\cite{mur2017orb} to handle the dynamic content with sparse point-level dynamic motion segmentation estimated by the correlation of their individual transformations across the temporal dimension.
Ye et al.~\cite{ye2022deflowslam} utilize dense optical flow and leverage a learning-based approach to estimate and mask out the content that moves independently of the rigid camera motion.
Building a VO pipeline for highly dynamic stereo surgical videos, Hayoz et al.~\cite{hayoz2023pose} combine background segmentation with an end-to-end learning-based method to estimate the contribution of each pixel's motion for camera pose optimization that is structured in a differentiable training architecture. 

We build \approach based on the findings of the prior work and utilize both the semantic segmentation~\cite{yu2018ds} of motion-prone objects and dense learning-based motion segmentation~\cite{ye2022deflowslam} to mask out dynamic content within a SLAM architecture~\cite{teed2021droid} to estimate camera pose.
Furthermore, we leverage this information with our introduced static-focused ray sampling method to further fine-tune the camera poses during neural reconstruction.

\subsection{Camera Localization and Radiance Fields}
Neural implicit methods gained popularity for scene representation.
Mildenhall et al.~\cite{mildenhall_nerf_2020} introduce NeRF employing a multi-layer perceptron (MLP) to represent a static 3D scene enabling NVS. NeRF requires precise camera poses during training, which are typically obtained through SfM pipelines like COLMAP~\cite{schonberger2016structure} that can take long computation times for large sets of input images.

Proposing an alternative solution that can jointly optimize camera poses and scene representation, Sucar et al.~\cite{sucar2021imap} reformulate NeRF in a SLAM framework for ordered sets of image sequences.
Zhu et al.~\cite{zhu2022nice} extend this by arranging multiple NeRFs in a hierarchical structure, improving the reconstruction capabilities. Following this, NICER-SLAM~\cite{zhu2023nicer} introduces the usage of estimated depth images for more accurate geometric reconstruction. Others~\cite{rosinol2022nerf,chung2023orbeez} utilize hash-based InstantNGP~\cite{muller2022instant} for its fast novel view rendering capabilities. NeRF-SLAM~\cite{rosinol2022nerf} adapts the Droid-SLAM~\cite{teed2021droid} framework and Orbeez-SLAM~\cite{chung2023orbeez} leverages ORB-SLAM~\cite{mur2017orb}.

More recent works~\cite{Matsuki_2024_CVPR,yan2024gs,zhu2024semgaussslamdensesemanticgaussian} utilize an explicit scene representation for static scenes, namely Gaussian Splatting (GS)~\cite{kerbl20233d} combined with SLAM architectures for even faster convergence and novel view rendering.

Unlike the prior work, our proposed method does not carry static-scene assumptions. To this end, we utilize a two-step approach that efficiently estimates the camera poses using a dynamics-aware SLAM method and represents the scene in a dynamic NeRF architecture~\cite{cao2023hexplane} that is extended to further refine the initially estimated camera poses.

\begin{figure*}[t!]
    \centering
    \includegraphics[width=\textwidth]{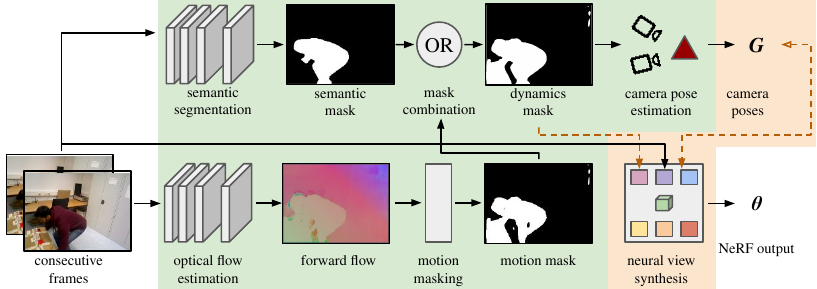}
    \caption{\approach retrieves consecutive RGB frames and applies semantic segmentation ($M_{ss} \rightarrow S_{S_i}$) and motion masks ($M_{MS} \rightarrow S_{M_i}$) on these input frames ($I$). Combining these masks (OR), we enable a motion-aware, fast and robust camera pose estimation $G$. Based on $G$, we produce a time-dependent output ($\Theta$) using dynamic NeRF. During NeRF training, we utilize the previously computed masks to refine the camera poses (dashed lines). Dashed lines represent iterative updates on NeRF and pose parameters. We highlight our approach in green.}
    \label{fig:archl}
\end{figure*}

\subsection{Neural Representation of Dynamic Scenes}

Our world is inherently dynamic which violates commonly used static scene assumptions and representations~\cite{zhu2022nice,zhu2023nicer,sucar2021imap,chung2023orbeez,rosinol2022nerf,muller2022instant,schieber2023nerftrinsic,mildenhall_nerf_2020}. To represent dynamics \cite{pumarola2021d,park2021nerfies}, the use of grid structures is common~\cite{cao2023hexplane,shao2023tensor4d,TiNeuVox}. 
HexPlane~\cite{cao2023hexplane} utilizes a 4D space-time grid divided into six feature planes. The feature vector is a 4D point in space-time projected onto each feature plane.
Similarly, Tensor4d~\cite{shao2023tensor4d} builds upon a 4D tensor field using time-aware volumes projected onto nine 2D planes.
TiNeuVox~\cite{TiNeuVox} represents dynamic scenes using time-aware voxel features to enable faster training.
~\cite{Wu_2024_CVPR,luiten2024dynamic} propose the usage of GS with point cloud initializations for dynamic scene representations.

Optimizing camera poses and view synthesis can improve results in static scenes~\cite{wang_nerf_2021,meng_gnerf_2021,yen-chen_inerf_2021,schieber2023nerftrinsic}.
Liu et al.~\cite{liu2023robust} introduce such an optimization for a dynamic NeRF using a static and a dynamic part.

We, in contrast, split the problem into the differentiation of static and dynamic content. Already at the initial camera pose estimation this split is considered. 
The dynamic environment is then represented by an efficient dynamic, 4D NeRF~\cite{cao2023hexplane} utilizing the camera poses estimated from the static scene components, and further refining these poses based on the static parts of the images.

\section{METHOD}

We utilize robust and fast camera localization and an implicit scene representation, a dynamic NeRF, to enable NVS on dynamic scenes. As input, we consider an RGB video captured in a dynamic environment using a single monocular moving camera without pose knowledge. This leads to two objectives: first, estimating the camera pose, and second, reconstructing the scene, which may comprise some deformation.
We leverage masks to separate the scene dynamics from the static background for the former and utilize a 4D spatio-temporal neural representation for the latter. This allows to optimize for NVS while refining initial camera poses.

\subsection{Robust Camera Localization in Dynamic Environments}

NVS methods commonly employ rigid SfM, often COLMAP~\cite{schonberger2016structure}, for camera pose estimation which assumes a static scene.
To cope with dynamics, it is possible to add binary masks to remove non-static regions. However, the high computation time of SfM approaches remains a drawback.
In our approach, we target both robustness against dynamic scenes and a more efficient solution for pose estimation for dynamic moving cameras.
Our underlying architecture is based on DROID-SLAM~\cite{teed2021droid}.
While it incorporates elements to cope with a certain level of scene dynamics, we found it can be improved by eliminating the contribution of the non-static regions for camera tracking.
We do this by generating binary dynamics masks for each frame, which are combinations of semantic (\(M_{SS}\)) and motion masks (\(M_{MS}\)).

The semantic mask follows the paradigm that in real-world videos, the most prominent source of motion is often an object of known class. We, therefore, define $M_{SS} = \mathbb{1}(x > 0) \circ \sum_i M_{SS}(i)$ for all defined semantic classes $i \in S$ of the set $S$ with the indicator function being $1$ for positive values.
Additional moving objects are covered through a motion mask $M_{MS}$. One example of a dynamic object outside of $S$ is depicted by the moving box in Figure~\ref{fig:archl}.
Following DytanVO~\cite{shen2023dytanvo}, our motion masks are generated using a CNN with the previous image, the forward optical flow and the estimated camera motion as input to iteratively refine the mask along with updates of the relative motion between frames.
Instead of random initialization, we begin with the motion mask of the previous frame to improve stability and lower the number of iterations in our refinement.
We combine both masks as $M = \mathbb{1}(x > 0) \left( M_{SS} + M_{MS} \right)$.

\subsection{Neural Dynamic Scene Representation}
To represent the dynamic scene, we build upon a state-of-the-art dynamic NeRF, HexPlane~\cite{cao2023hexplane}.
Targeting efficient optimization, fast rendering and high reconstruction quality, HexPlane~\cite{cao2023hexplane} utilizes TensoRF~\cite{chen2022tensorf} and constructs a 4D representation with three spatial and one temporal axes to create feature planes. Their pairwise combinations are then the input of a tiny MLP.
The grid size of each axis is governed in a coarse-to-fine fashion along the training to improve convergence and naturally regularizes the local neighborhoods.
For optimization, we follow the original NeRF ray casting and apply the pixel-wise photometric loss between the rendered RGB values $C$ and the ground truth values $\hat{C}$, supported by a total variation regularization as:
\begin{equation}
\label{eq:loss}
\mathcal{L}=\frac{1}{|\mathcal{R}|} \sum_{\mathbf{r} \in \mathcal{R}}\|C(\mathbf{r})-\hat{C}(\mathbf{r})\|_2^2
+ \lambda_{\text {TV}} \mathcal{L}_{\text {TV}}.
\end{equation}
While $\mathcal{R}$ represents the sampled rays, $\mathcal{L}_{\text{TV}}$ and $\lambda_{\text{TV}}$ depict the total variation regularization and its weight.

\subsection{Pose Refinement and Training Strategy}

We utilize eqn.~\eqref{eq:loss} as the objective function for training both the NeRF and refining the camera poses. During optimization, we employ an alternating training strategy in which every 5\,000 iterations, the parameters to be optimized are switched between the radiance field and the pose parameters.
To improve the accuracy and robustness of the optimization, during the pose refinement phase, we constrain the sampled rays, $\mathcal{R}$, contributing to the loss to be elements of the static regions designated by the inverse dynamics masks.
Furthermore, we also use the same masks to ensure that the total variation regularization is imposed only on the static components.
As the amount of non-dynamic sampled rays can vary from image to image, we scale the learning rate $\eta_s$ by the fraction of static pixels in the image to keep the optimization steps in the same magnitude:
\begin{equation}
    \eta = \eta_s * \frac{n_{static}}{n_{static} + n_{dynamic}},
\end{equation}
where $n_{static}$ and $n_{dynamic}$ represent the number of static and dynamic pixels in the image and $\eta$ is the learning rate used for the pose refinement.

\section{EVALUATION}

To evaluate our approach, we consider the translational absolute trajectory error (ATE)~\cite{zhang2018tutorial} using root mean square error (RMSE) in meters. To reason about the orientation, we employ the Relative Rotation Accuracy (RRA) as defined in \cite{wang2023posediffusion} after synchronizing the poses to the GT to be invariant against the alignment of coordinate frames. We calculate the median RRA over all sequences and report the percentage of poses for which the RRA lies below a specified threshold. For NVS quality, we analyze the rendering quality of the novel dynamic views by reporting PSNR and SSIM~\cite{wang_image_2004}.

\begin{table}[t!]
\centering
\caption{
Percentage of orientations for which the median RRA is below the specified threshold [\%] on the TUM RGB-D dataset and the Bonn RGB-D dynamic dataset. \label{tab:boxplot}
}
    \resizebox{\columnwidth}{!}{ 
    \begin{tabular}{l|l|cccc} \hline\hline
        Dataset & Metric & DROID-SLAM & OURS (MS\&SS) & OURS (SS) & OURS (MS)\\\hline
        %\textit{TUM RGB-D} & \\
        TUM RGB-D & median RRA@5 $\uparrow$ & \textbf{100.0} & \textbf{100.0} & \textbf{100.0} & \textbf{100.0}\\
        & median RRA@15 $\uparrow$ & \textbf{100.0} & \textbf{100.0} & \textbf{100.0} & \textbf{100.0}\\ \hline
        %\textit{Bonn RGB-D Dynamic} & \\
        Bonn RGB-D & median RRA@5 $\uparrow$ &  61.05 & \textbf{61.89} & 61.86 & 61.88\\
        & median RRA@15 $\uparrow$ &  91.29 & 91.45 & \textbf{91.46} & 91.27\\\hline\hline
    \end{tabular}}
\label{tab:RRA}
\end{table}

\subsection{Datasets}

To assess \approach, we evaluate both the camera localization and the NVS part with state-of-the-art approaches on two dynamic datasets. In total, we evaluate our approach on 33 dynamic sequences (9 from TUM-RGB-D, and 24 from BONN RGB-D). For NVS, we follow the usual splitting using every eighth frame for testing. 

\paragraph{\textbf{TUM RGB-D - Dynamic subset}~\cite{sturm12iros}}. We used the nine dynamic sequences,  with five sequences rated \emph{slightly dynamic} and four \emph{highly dynamic}. The images of the TUM RGB-D dataset have a resolution of $640\times480$. 

\paragraph{\textbf{BONN RGB-D Dynamic}~\cite{palazzolo2019iros}} The BONN RGB-D Dynamic dataset consists of 24 dynamic sequences and two static sequences with an image resolution of $640\times480$.

\begin{table*}[t!]
\renewcommand{\arraystretch}{1.1}
    \caption{RMSE of the translational ATE [m] on TUM RGB-D~\cite{sturm12iros}. The upper rows are the slightly dynamic scenes and the lower rows are the highly dynamic scenes. The - denotes that no results were reported for these sequences.}
    \label{tab:slam_tum}
    \begin{center}
    \resizebox{\textwidth}{!}{
    \begin{tabular}{lcccccccccccc}  \hline \hline  
        \multirow{2}{*}{Sequences} & & & RGB-D & & & & \multicolumn{6}{c}{RGB}  \\ \cline{2-6}  \cline{8-13}
         & DVO-SLAM & ORB-SLAM2 & PointCorr & DynaPix-D~\cite{xu2023dynapix} & Yu et al. & & DytanVO & DeFlowSLAM & DROID-SLAM & OURS (MS\&SS) & OURS (SS) & OURS (MS)\\
         & \cite{kerl2013robust} &\cite{mur2017orb} & \cite{dai2020rgb} & \cite{xu2023dynapix} &\cite{Yu2024RobustVS} & & \cite{shen2023dytanvo} & \cite{ye2022deflowslam}  &\cite{teed2021droid} & & & \\\hline
         fr2/desk-person & 0.104 & \textbf{0.006} & 0.008 & - & - & &  1.166 & 0.013 & \textbf{0.006} & 0.007 & 0.007 & \textbf{0.006}\\
         fr3/sitting-static & 0.012 & 0.008 & 0.010 &- & 0.006 &&  0.016 & 0.007 & \textbf{0.005} & \textbf{0.005} & \textbf{0.005} & \textbf{0.005}\\
         fr3/sitting-xyz & 0.242 & 0.010 & \textbf{0.009} & -&-&&  0.260 & 0.015 & \textbf{0.009} & 0.010 & 0.010 & \textbf{0.009}\\
         fr3/sitting-rpy & 0.176 & 0.025 & 0.023 &-&-&&  0.046 &  0.027 & 0.022 & 0.024 & 0.024 & \textbf{0.021}\\
         fr3/sitting-halfsphere & 0.220 & 0.025 & 0.024 &-& 0.018 & & 0.310 & 0.025 & \textbf{0.014} & 0.023 & 0.028 & 0.019\\ \hline
        fr3/walking-static & 0.752 & 0.408 &  0.011 & \textbf{0.007} & 0.009 & & 0.021 & \textbf{0.007} & 0.012 & \textbf{0.007} & \textbf{0.007} & 0.014\\ 
         fr3/walking-xyz & 1.383 & 0.722 & 0.087 & \textbf{0.014} & \textbf{0.014} & & 0.028 & 0.018 & 0.016 & \textbf{0.014} & \textbf{0.014} & \textbf{0.014}\\ 
         fr3/walking-rpy & 1.292 & 0.805 & 0.161 & 0.123 & \textbf{0.031} & & 0.155 & 0.057 & 0.040 & \textbf{0.031} & 0.036 & 0.039\\ 
         fr3/walking-halfsphere & 1.014 & 0.723 & 0.035 & 0.023 & 0.020 &&  0.385 & 0.420 & 0.022 & \textbf{0.019} & \textbf{0.019} & 0.020\\ \hline
         Mean/Max & 0.577/1.383 & 0.304/0.805 & 0.041/0.161 &-&-&&  0.265/1.166 & 0.065/0.420 & \textbf{0.016}/0.040 & \textbf{0.016}/\textbf{0.031} & 0.017/0.036 & \textbf{0.016}/0.039\\ \hline\hline
    \end{tabular}
    }
    
    \end{center}
\end{table*}

\begin{table*}[t!]
\renewcommand{\arraystretch}{1.1}
        \caption{RMSE of the translational ATE [m] on BONN RGB-D Dynamic~\cite{palazzolo2019iros}. RGB-D results are reported by Palazzolo et al.~\cite{palazzolo2019iros}. For DROID-SLAM and ours, we calculated the results.}
    \label{tab:slam_bonn}
    \begin{center}
    \resizebox{\textwidth}{!}{
    \begin{tabular}{lccccccccc}  \hline \hline  
                      \multirow{2}{*}{Sequences} & \multicolumn{4}{c}{RGB-D} & &\multicolumn{4}{c}{RGB}  \\ \cline{2-5}  \cline{7-10}
           & ReFusion~\cite{palazzolo2019iros} & StaticFusion~\cite{scona2018staticfusion} & DynaSLAM (G)~\cite{bescos2018dynaslam} & DynaSLAM (N+G)~\cite{bescos2018dynaslam} & & DROID-SLAM~\cite{teed2021droid} & OURS (MS\&SS) & OURS (SS) & OURS (MS)\\ \hline
         balloon & 0.175 & 0.233 & 0.050 & 0.030 & & 0.075 & \textbf{0.028} &  0.030 & 0.068\\
         balloon2 & 0.254 &  0.293 & 0.142 & 0.029 & & 0.041 & \textbf{0.027} & \textbf{0.027} & 0.038\\
         balloon\_tracking & 0.302 & 0.221 & 0.156 & 0.049 & & 0.035 & \textbf{0.034} & 0.037 & 0.036\\
         balloon\_tracking2 & 0.322 & 0.366 & 0.192& 0.035 &&  \textbf{0.026} & 0.032 & 0.029 & 0.028\\
         crowd & 0.204 & 3.586 & 1.065 & \textbf{0.016} &&  0.052 & 0.035 & 0.033 & 0.061\\
         crowd2 & 0.155 & 0.215 & 1.217 & 0.031 &&  0.065 & \textbf{0.028} & 0.029 & 0.056\\
         crowd3 & 0.137 & 0.168 & 0.835 & 0.038 &&  0.046 & \textbf{0.032} & 0.052 & 0.055\\  
         kidnapping\_box & 0.148 & 0.336 & 0.026 &0.029 & & \textbf{0.020} & 0.021 & \textbf{0.020} & \textbf{0.020}\\  
         kidnapping\_box2 &  0.161 & 0.263 & 0.033 & 0.035 &&  \textbf{0.017} & \textbf{0.017} & \textbf{0.017} & \textbf{0.017}\\  
         moving\_no\_box & 0.071 & 0.141 &0.317& 0.232 &&  0.023 & \textbf{0.013} & 0.014 & 0.014\\  
         moving\_no\_box2 & 0.179 &  0.364 & 0.052 & 0.039 & & 0.040 & 0.027 & 0.027 & \textbf{0.026} \\  
         moving\_o\_box & 0.343 & 0.331 & 0.544 & \textbf{0.044} &&  0.177 & 0.152 & 0.152 & 0.167\\  
         moving\_o\_box2 & 0.528 & 0.309 & 0.589 & 0.263 & & 0.236 & \textbf{0.175} & 0.176 & 0.176\\  
         person\_tracking & 0.289 & 0.484 & 0.714 & 0.061 &&  0.043 & 0.148 & 0.033 & \textbf{0.024}\\  
         person\_tracking2 & 0.463 & 0.626 & 0.817 & 0.078 & & 0.054 & \textbf{0.022} & 0.023 & 0.035\\  
         placing\_no\_box & 0.106 & 0.125 & 0.645 & 0.575 &&  0.078 & 0.021 & 0.042 & \textbf{0.018}\\  
         placing\_no\_box2 & 0.141 & 0.177 & 0.027 & 0.021 &&  0.030 & 0.020 & \textbf{0.019} & 0.020\\  
         placing\_no\_box3 & 0.174 & 0.256 & 0.327 & 0.058 &&  0.025 & \textbf{0.022} & \textbf{0.022} & \textbf{0.022}\\  
         placing\_o\_box & 0.571 & 0.330 & 0.267 & 0.255 & & 0.127 & 0.172 & 0.173 & \textbf{0.117}\\  
         removing\_no\_box &  0.041 & 0.136 &  0.016 &  0.016 & & 0.016 & \textbf{0.015} & \textbf{0.015} & \textbf{0.015}\\  
         removing\_no\_box2 & 0.111 & 0.129&  0.022 & 0.021 &&  0.020 & \textbf{0.019} & 0.020 & 0.021\\  
         removing\_o\_box & 0.222 & 0.334&  0.362 & 0.291 & & 0.189 & \textbf{0.177} & 0.179 & 0.181\\  
         synchronous & 0.441 & 0.446 & 0.977&  0.015 &&  \textbf{0.006} & 0.007 & 0.007 & \textbf{0.006}\\  
         synchronous2 & 0.022 & 0.027&  0.887 & 0.009 &&  0.012 & \textbf{0.006} & 0.007 & 0.011\\  \hline
         Mean/Max & 0.232/0.571 & 0.412/3.586 &  0.428/1.217 & 0.095/0.575 & & 0.061/0.236 & 0.052/\textbf{0.177} & \textbf{0.049}/0.179 & 0.051/0.181\\ \hline\hline
    \end{tabular}
    }
   \end{center}
\end{table*}

\subsection{Implementation Details}
We implemented our approach in Python using PyTorch and evaluated it on one single workstation (Intel i9-10980XE CPU, NVIDIA GeForce RTX 3090 GPU 24GB, and 128GB RAM) unless otherwise specified.

For semantic masking, we use DeepLabV3 \cite{chen2017rethinking}/ResNet50 \cite{he2016deep} trained on COCO \cite{lin2014microsoft} to obtain human masks. For motion masking, we apply DytanVO~\cite{shen2023dytanvo}. In practice, initialized with the motion mask of the previous frame, we found that setting the number of iterations for motion mask refinement to two provides stable convergence.

When the camera motion exceeds that of the DytanVO training sequences, the motion segmentation module produces more false positives. To counter this, we set a threshold of 60\% for the maximum number of motion pixels; masks exceeding this threshold are discarded to maintain sufficient pixels for dense bundle adjustment. Additionally, a pixel is considered dynamic only if the confidence is above 0.95 for the first mask refinement step and 0.98 for subsequent steps.

For NVS, we use the full image size. During camera localization, the images are downsampled by half to remain consistent with existing trajectory experiments~\cite{teed2021droid}.

We set the loss parameter $\lambda_{TV}$ to 0.005 and sample $\lvert\mathcal{R}\rvert = 1,024$ rays per batch. We use the default optimization settings for the dynamic NeRF~\cite{cao2023hexplane}, with a multi-step learning rate decay scheduler for pose refinement. The initial learning rate is $1 \times 10^{-5}$, decayed by a factor of 0.9 every $1,000$ iterations and reset in each pose optimization cycle. The learning rate is scaled by the percentage of static pixels in the image. The number of optimization iterations is set to $200,000$, unless specified otherwise.

\subsection{Camera Localization Quality}

Our results on all datasets show a lower translational trajectory error compared to state-of-the-art approaches, see Table~\ref{tab:slam_tum} and Table~\ref{tab:slam_bonn}. We compare RGB-D-based and monocular SLAM approaches with our proposed camera retrieval method using only motion masks (MS), using only semantic masks (SS) as well as using both types of masks in combination (MS\&SS). Looking at the mean value of the RMSE of the translational ATE over all sequences, our approach performs similarly to DROID-SLAM on TUM RGB-D. However, the maximum error value is much lower, which shows that it is less prone to failure when applied to real-world scenarios. On the BONN RGB-D Dynamic dataset, our approach outperforms the state-of-the-art approaches in terms of mean and maximum trajectory error for translation. Moreover, comparing the RRA, see Table~\ref{tab:RRA}, our MS\&SS-model achieves higher values than DROID-SLAM on the BONN RGB-D dataset, while all methods perform equally well on TUM RGB-D. It can be noted that using one type of mask is usually sufficient to improve over DROID-SLAM. In some cases, like the person-tracking scene, where the combination performs worse than SS or MS, the SLAM backbone seems to have too less pixels left for the bundle adjustment.

\subsection{Novel View Synthesis Quality}

\begin{figure*}[t!]
    \centering
    \includegraphics[width=\textwidth]{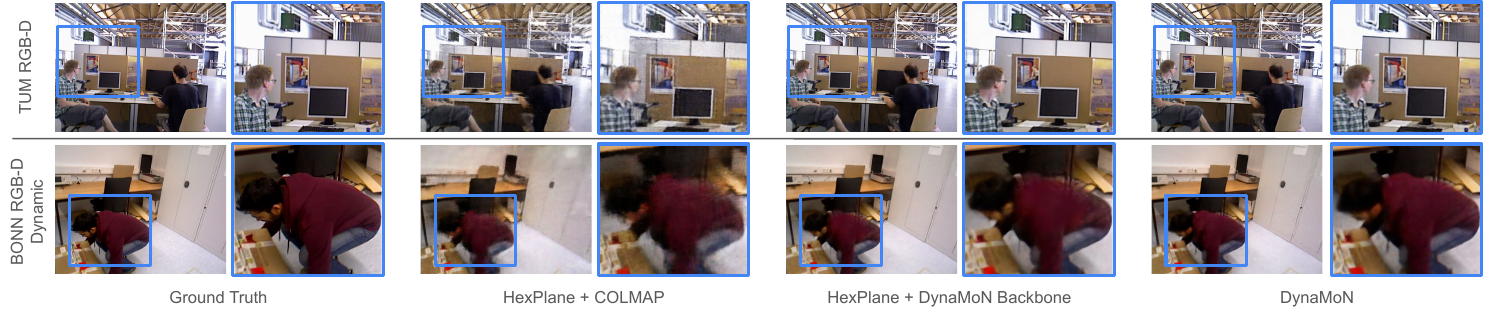}
    \caption{Qualitative Results for NVS on TUM RGB-D (top) and BONN RGB-D Dynamic  (bottom). Ground Truth (left),  novel views views from HexPlane + COLMAP (2nd column), improved camera poses from our backbone (3rd column), and our full model (right). \label{fig:example_tum}}
    
\end{figure*}

To evaluate the NVS quality of our approach, we consider several aspects. First, we compare it with the traditional way of generating camera ground truth for 4D NeRF~\cite{cao2023hexplane} and 4D GS~\cite{Wu_2024_CVPR} using COLMAP, see Tables~\ref{tab:tum_nerf} and~\ref{tab:bonn_nerf}. We follow the implementation of InstantNGP~\cite{muller2022instant} to retrieve the COLMAP ground truth.
Second, we compare our approach with RoDynRF~\cite{liu2023robust} without initial COLMAP poses, shown in Table~\ref{tab:tum_nerf}. We follow RoDynRF's training guidelines.
Third, we highlight the influence of the iterative pose refinement by training both \approach and HexPlane with our initially localized poses, see Tables ~\ref{tab:tum_nerf}, Table~\ref{tab:bonn_nerf} and ~\ref{tab:ablation}. Qualitative examples of renderings on both datasets can be found in Figure~\ref{fig:example_tum}.

On both datasets, COLMAP is challenged by the dynamics, and training with COLMAP camera poses can only be performed on a subset of scenes. \approach retrieves all camera poses for both datasets, see Tables~\ref{tab:slam_tum} and~\ref{tab:slam_bonn}. As visualized in Figure~\ref{fig:example_tum}, the visual results of COLMAP-based NeRF are noisy on both datasets. \approach's camera poses can show improved results. The complete \approach can further improve details. In terms of PSNR/SSIM, \approach leads to a higher NVS quality compared to HexPlane without refined poses, RoDynRF, and when using COLMAP for the camera poses for either 4D NeRF or 4D GS, see Table~\ref{tab:tum_nerf}. 

\begin{table*}[t!]
    \caption{{NVS results on TUM RGB-D}. The upper rows are the slightly dynamic scenes and the lower rows are the highly dynamic scenes. The - denotes that the camera pose could not be regressed. The * denotes random point cloud initialization, as the one from COLMAP lead to no convergence. We report PSNR/SSIM.}
    \label{tab:tum_nerf}
   \begin{center}
   %\resizebox{\columnwidth}{!}{ 
    \resizebox{0.7\textwidth}{!}{ 
    \begin{tabular}{lcccccc} \hline \hline
      \multirow{2}{*}{PSNR$\uparrow$/SSIM$\uparrow$} & \multicolumn{1}{c}{RGB-D} & & \multicolumn{4}{c}{RGB}  \\ \cline{2-2}  \cline{4-7} 
       \multirow{2}{*}{Sequences} & RoDynRF~\cite{liu2023robust} &  & \specialcell{4D Gaussians~\cite{Wu_2024_CVPR} \\ (COLMAP)} & \specialcell{HexPlane~\cite{cao2023hexplane} \\ (COLMAP)} & \specialcell{HexPlane \\+ \approach \\ Backbone (Ours)} & \specialcell{\approach \\ (Ours)} \\  \hline 
       fr2/desk-person & 6.05 / 0.127 &  & - & - & 23.04 / 0.680 & \textbf{24.83 / 0.725}\\
       fr3/sitting-static & 9.67 / 0.123 & & - & - & 28.55 / 0.899 & \textbf{28.95 / 0.903}\\
       fr3/sitting-xyz & 10.10 / 0.150 & & \textbf{26.40 / 0.886} &  19.98 / 0.589 & {25.90 / 0.848} & 26.15 / 0.850\\
       fr3/sitting-rpy & 8.35 / 0.121  &  & 24.15 / 0.814 & 16.63 / 0.472 & 25.76 / 0.843 & \textbf{26.40 / 0.853}\\
       fr3/sitting-halfsphere &  7.46 / 0.157 &  & 24.42 / \textbf{0.830} & 24.25 / 0.793  &  24.05 / 0.792 & \textbf{24.60} / 0.799\\ \hline 
       fr3/walking-static   &  10.01 / 0.159 &  & 16.39 / 0.575* & 20.27 / 0.650 & {25.84 / 0.849} & \textbf{26.25 / 0.853}\\ 
       fr3/walking-xyz &  10.02 / 0.171 &  & - & - & 24.05 / \textbf{0.797} & \textbf{24.34} / 0.796 \\ 
        fr3/walking-rpy & 8.81 / 0.201 &  & - & - & {24.18 / 0.790} & \textbf{24.70 / 0.796}\\ 
        fr3/walking-halfsphere  & 8.79 / 0.205 &  &  21.72 / 0.746 & 22.01 / 0.692 & 24.05 / \textbf{0.781} & \textbf{24.27} / 0.779\\ \hline
         Mean &8.84 / 0.157 & & - & - & {25.05 / 0.809} & \textbf{25.60 / 0.819}\\ \hline\hline
    \end{tabular}
   }
   \end{center}
\end{table*}

\begin{table}[t!]
    \caption{{NVS results on BONN RGB-D Dynamic.} The - denotes that the camera pose could not be regressed. We report PSNR/SSIM.}
    \label{tab:bonn_nerf}
   \begin{center}
     \resizebox{\columnwidth}{!}{ 
    \begin{tabular}{lccc} \hline \hline
       \multirow{2}{*}{PSNR$\uparrow$/SSIM$\uparrow$} & \multicolumn{3}{c}{RGB}  \\ \cline{2-4} 
       \multirow{2}{*}{Sequences}  & \specialcell{HexPlane \\ (COLMAP)} & \specialcell{HexPlane +\\ \approach \\ Backbone (Ours)} & \specialcell{ \approach \\ (Ours)} \\  \hline 
        balloon & - &  29.83 / \textbf{0.868} & \textbf{29.84} / 0.866\\
         balloon2 & - &  27.61 / 0.841 & \textbf{28.31 / 0.847}\\
         balloon\_tracking & - &  29.27 / \textbf{0.847}& \textbf{29.55} / 0.846\\
         balloon\_tracking2 & - &  29.68 / 0.860& \textbf{30.56 / 0.862}\\
         crowd & - &  27.92 / \textbf{0.831}& \textbf{28.40} / 0.829\\
         crowd2 & - &  26.68 / 0.825& \textbf{27.97 / 0.827}\\
         crowd3 & - &  26.57 / 0.815 & \textbf{27.40 / 0.827}\\
         kidnapping\_box & - &  30.43 / \textbf{0.870} & \textbf{30.59} / 0.869\\
         kidnapping\_box2 & \textbf{30.87 / 0.881} & 30.28 / 0.860 & 30.18 / 0.857\\
         moving\_no\_box &  \textbf{30.67 / 0.881} & 30.32 / 0.866 & 30.36 / 0.862\\ 
         moving\_no\_box2 &  \textbf{31.14 / 0.890} & 30.57 / 0.874 & 30.62 / 0.870\\ 
         moving\_o\_box & 29.21 / \textbf{0.889} &  30.26 / 0.884 & \textbf{31.08} / 0.888\\  
         moving\_o\_box2 & - &  \textbf{29.82 / 0.875} & 29.40 / \textbf{0.875}\\
         person\_tracking & 26.57 / 0.808 &  27.78 / 0.843& \textbf{28.71 / 0.845}\\
         person\_tracking2 & - &  27.47 / \textbf{0.837}& \textbf{27.80} / 0.833\\
         placing\_no\_box & 25.73 / 0.780 &  30.94 / 0.869& \textbf{31.23 / 0.870}\\
         placing\_no\_box2 & - & 31.18 / \textbf{0.885}& \textbf{31.45} / 0.884\\ 
         placing\_no\_box3 & - & 30.79 / \textbf{0.879}& \textbf{30.83} / 0.875\\  
         placing\_o\_box & - & 30.42 / 0.886& \textbf{31.12 / 0.890}\\
         removing\_no\_box & - & 30.66 / \textbf{0.879} & \textbf{30.85} / 0.877\\
         removing\_no\_box2 & \textbf{31.13 / 0.887} & 30.73 / 0.871& 30.86 / 0.870\\
         removing\_o\_box & \textbf{31.13 / 0.887} & 30.30 / 0.880& 30.76 / 0.882\\
         synchronous & - & 28.42 / \textbf{0.835}& \textbf{28.93} / 0.834\\ 
         synchronous2 & - & 28.85 / 0.887 & \textbf{30.18 / 0.891}\\ \hline
         Mean & - & 29.45 / \textbf{0.861} & \textbf{29.87 / 0.861}\\ \hline\hline
    \end{tabular}
    }
   \end{center}
\end{table}

\begin{table}[t!]
    \centering
    \caption{
    Analysis of different pose retrieval steps on NVS quality, evaluated on the crowd2 scene of the BONN RGB-D Dynamic dataset. We assess robustness using a sparse input trajectory (every 10th frame).
    }
    \resizebox{0.7\columnwidth}{!}{ 
    \begin{tabular}{l|c} \hline\hline
    Approach & \specialcell{PSNR$\uparrow$/SSIM$\uparrow$}\\\hline
    DROID-SLAM~\cite{teed2021droid} + HexPlane~\cite{cao2023hexplane} & 16.46 / \textbf{0.661}\\
    Our initial poses + HexPlane~\cite{cao2023hexplane} & 17.07 / 0.660\\\hline
    DynaMoN (Ours) & \textbf{17.31} / 0.659\\\hline\hline
    \end{tabular}
    }
    \label{tab:ablation}
\end{table}

\subsection{Robustness on Sparse Trajectories}

Dynamic scenes pose more challenges regarding the robustness of an approach compared to static environments. In  Table~\ref{tab:ablation}, we showcase the robustness of our approach. We use only every tenth frame of the crowd2 sequence from the BONN RGB-D Dynamic dataset. As reported in  Table~\ref{tab:ablation}, we show how the underlying camera localization approach influences the NVS result. When comparing DROID-SLAM camera poses to the ones from our \approach backbone and our neural camera pose refinement, the best NVS result can be achieved using \approach. Looking at the PSNR values in Table~\ref{tab:ablation}, it can be noted that each of our proposed components increases the reconstruction quality. Thus, when confronted with a sparse trajectory of a highly dynamic scene, \approach outperforms pure DROID-SLAM making it more robust to realistic, real-world scenarios with imperfect camera recordings.

\subsection{Runtime and Memory Evaluation}

We compare the runtime and memory consumption of \approach with RoDynRF~\cite{liu2023robust} on a workstation with an NVIDIA RTX 4090 graphics card with 24 GB of VRAM. We report the results for the TUM RGB-D fr2/desk-person sequence in Table~\ref{tab:runtime} as it is by far the longest scene of both datasets comprising 4\,067 frames. Using the COLMAP configuration from Instant-NGP, COLMAP runs for 38:28:16 (hours:minutes:seconds), see Table~\ref{tab:runtime}.
In terms of memory requirement, RoDynRF can only be trained with a batch size of 512 in this setting, while \approach is more memory-efficient, enabling a batch size of 1\,024. From the experiments, it can be concluded that the total time for pre-processing, training and rendering of \approach is comparable to RoDynRF if the default training strategy with 200\,000 iterations is utilized. However, if our model is trained for 100\,000 iterations which is the default value of RoDynRF, the total time can be greatly reduced without a significant drop in photometric accuracy. For this shorter training, the NeRF and pose optimization of \approach alternate every 2\,500 iterations. Furthermore, a decrease of the batch size to the same value as in the competitor method, RoDynRF, further speeds up the training time at the cost of a longer rendering time while staying comparably accurate.

It has to be noted that the pre-processing of RoDynRF consisting of monocular depth prediction, optical flow prediction, motion mask generation and dataloading takes about eight times longer than our pre-processing loading the data and generating initial poses as well as motion masks.

\begin{table}[t!]
\centering
\caption{
Analysis of the runtime on TUM RGB-D fr2/desk-person. Ours refers to 200,000 iterations / 1,024 batch size. The symbol $\ddagger$ indicates 100,000 iterations / 512 batch size, while $\dagger$ represents 100,000 iterations / 1.024 batch size. Time is given in the format hours:minutes:seconds
}
    \resizebox{\columnwidth}{!}{ 
    \begin{tabular}{l|ccc|c|c} \hline\hline
        Approach & Pre-processing$\downarrow$ & Training$\downarrow$ & Rendering$\downarrow$ & Total$\downarrow$ & PSNR$\uparrow$/SSIM$\uparrow$\\\hline
        COLMAP~\cite{schonberger2016structure} & 38:28:16 & - & - & - & - \\ \hline 
        RoDynRF~\cite{liu2023robust} $\ddagger$ & 01:52:55 & 08:00:44 & 00:11:15 & 10:04:54 & 6.05 / 0.127 \\\hline
        Ours & 00:13:49 & 09:49:25 & 00:42:32 & 10:45:46 & 24.83 / 0.725\\
        Ours $\dagger$ & 00:14:12 & 05:04:19 & 00:42:20 & 06:00:51 & 23.95 / 0.702\\
        Ours $\ddagger$ & 00:13:41 & 04:43:23 & 01:03:21 & 06:00:25 & 23.43 / 0.690\\\hline\hline
    \end{tabular}}
\label{tab:runtime}
\end{table}
    
\section{DISCUSSION}

In comparison to traditional sparse reconstructions~\cite{campos2021orb,mur2015orb,teed2021droid}, our approach addresses camera localization and dynamic NVS in combination. While this leads to more pleasing 3D visualizations and NVS results, the computing time is higher compared to traditional camera localization and non-dynamic 3D representation methods~\cite{zhu2022nice, zhu2023nicer,chung2023orbeez}. Nevertheless, compared to combined dynamic camera localization and dynamic NVS approaches~\cite{liu2023robust}, our approach requires only 60\% of the training time when using the same number of iterations and batch size, while being more robust in more dynamic scenes with dynamic camera motion.

Moreover, our approach shows an improved performance compared to the state-of-the-art for camera localization. 
Only using our motion masks already improves the performance. Moreover, for scenes with large motions, the combination with semantic segmentation masks provides significantly improved camera localization results. The results presented in Table~\ref{tab:tum_nerf} indicate that RoDynRF~\cite{liu2023robust}, despite being a state-of-the-art method that integrates combined camera regression with NVS, faces significant challenges in handling highly dynamic camera movements and dynamic scenes.
In contrast, our method demonstrates superior robustness and yields improved outcomes when compared to both RoDynRF~\cite{liu2023robust} and HexPlane~\cite{cao2023hexplane} employing COLMAP poses on the BONN RGB-D Dynamic and TUM RGB-D dataset.
Furthermore, our iterative pose refinement process enhances robustness, as evidenced in Table~\ref{tab:ablation}, and achieves better performance in NVS on both datasets.

Looking at the results of 4D GS with COLMAP poses, combining robust camera localization and 4D GS seems to be a promising future research direction.

\section{LIMITATIONS}

\approach is constrained by the scene representation capabilities of dynamic NeRFs. Although it performs well on large-scale datasets with up to over $4,000$ images, handling scenes with close to $8,000$ frames introduces significant challenges to the camera pose localization~\cite{li2024rd} and the NeRF~\cite{tancik2022block}. Nevertheless, our method works effectively with datasets like TUM RGB-D and BONN RGB-D Dynamic. Still, larger datasets~\cite{cortes2018advio} pose challenges, for NeRF and the camera localization~\cite{li2024rd}. The underlying dynamic NeRF in this work has difficulties representing environments of this scale, limiting its applicability to more extensive datasets.

\section{CONCLUSION}

We present \approach, a motion-aware, fast and robust camera localization approach for NVS. \approach can cope with the motion of objects of known and unknown classes using the combined segmentation and motion mask resulting in the dynamics mask. \approach retrieves camera poses faster and is more robust to scene dynamics in comparison to traditional SfM approaches, enabling a more accurate 4D scene representation. In addition, \approach is trained with a novel iterative training scheme that refines NeRF parameters and the initially provided poses in an alternating fashion. Compared to the state-of-the-art, \approach outperforms other dynamic camera localization approaches and shows better results for NVS on scenes with high camera motion.

%%%%%%%%%%%%%%%%%%%%%%%%%%%%%%%%%%%%%%%%%%%%%%%%%%%%%%%%%%%%%%%%%%%%%%%%%%%%%%%%

%%%%%%%%%%%%%%%%%%%%%%%%%%%%%%%%%%%%%%%%%%%%%%%%%%%%%%%%%%%%%%%%%%%%%%%%%%%%%%%%

%%%%%%%%%%%%%%%%%%%%%%%%%%%%%%%%%%%%%%%%%%%%%%%%%%%%%%%%%%%%%%%%%%%%%%%%%%%%%%%%

\bibliographystyle{IEEEtran} % We choose the "IEEEtran" reference style
\bibliography{IEEEabrv, reflist} % Entries are in the refs.bib file

\end{document}